\documentclass{article}

\usepackage{times}
\usepackage{graphicx} 
\usepackage{amsmath} 
\usepackage{amssymb} 
\usepackage{caption}
\usepackage{subcaption}
\usepackage[export]{adjustbox}

\usepackage{natbib}
\usepackage{multirow}

\usepackage{algorithm}
\usepackage{algorithmic}

\usepackage{booktabs}

\usepackage{hyperref}

\renewcommand{\vec}{\mathbf}

\usepackage[accepted]{icml2016}

\icmltitlerunning{Pixel Recurrent Neural Networks}

\begin{document} 

\twocolumn[
\icmltitle{Pixel Recurrent Neural Networks}

\icmlauthor{A\"aron van den Oord}{avdnoord@google.com}
\icmlauthor{Nal Kalchbrenner}{nalk@google.com}
\icmlauthor{Koray Kavukcuoglu}{korayk@google.com}
\icmladdress{\vspace{0.3cm} Google DeepMind}

\icmlkeywords{PixelRNN, Discrete Image Distribution, Image Density Models, Recurrent Neural Networks}

\vskip 0.3in
]

\begin{abstract} 

Modeling the distribution of natural images is a landmark problem in unsupervised learning.
This task requires an image model that is at once expressive, tractable and scalable. We present a deep neural network that sequentially predicts the pixels in an image along the two spatial dimensions. Our method models the discrete probability of the raw pixel values and encodes the complete set of dependencies in the image. Architectural novelties include fast two-dimensional recurrent layers and an effective use of residual connections in deep recurrent networks. 
We achieve log-likelihood scores on natural images that are considerably better than the previous state of the art.
Our main results also provide benchmarks on the diverse ImageNet dataset. Samples generated from the model appear crisp, varied and globally coherent.

\end{abstract} 

\section{Introduction}

\begin{figure}[!t]
\centering
\small
\hspace{0.02cm} {occluded} \hfill completions \hfill{original} \,

\vspace{0.1cm}
\includegraphics[ width=0.98\linewidth]{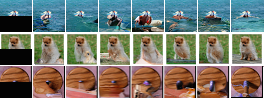} 
\vspace{-0.3cm}
\caption{Image completions sampled from a PixelRNN.}
\label{fig:intro_completions}
\vspace{-0.5cm}
\end{figure}

Generative image modeling is a central problem in unsupervised learning. Probabilistic density models can be used for a wide variety of tasks that range from image compression and forms of reconstruction such as image inpainting (e.g., see Figure \ref{fig:intro_completions}) and deblurring, to generation of new images. When the model is conditioned on external information, possible applications also include creating images based on text descriptions or simulating future frames in a planning task. One of the great advantages in generative modeling is that there are practically endless amounts of image data available to learn from. However, because images are high dimensional and highly structured, estimating the distribution of natural images is extremely challenging.

One of the most important obstacles in generative modeling is building complex and expressive models that are also tractable and scalable. This trade-off has resulted in a large variety of generative models, each having their advantages. Most work focuses on stochastic latent variable models such as VAE's \cite{rezende2014stochastic, DBLP:journals/corr/KingmaW13} that aim to extract meaningful representations, but often come with an intractable inference step that can hinder their performance.

One effective approach to \emph{tractably} model a joint distribution of the pixels in the image is to cast it as a product of conditional distributions; this approach has been adopted in autoregressive models such as NADE \cite{larochelle2011} and fully visible neural networks \cite{neal1992connectionist, Bengio_Bengio_NIPS99}. The factorization turns the joint modeling problem into a sequence problem, where one learns to predict the next pixel given all the previously generated pixels. But to model the highly nonlinear and long-range correlations between pixels and the complex conditional distributions that result, a highly expressive sequence model is necessary.

Recurrent Neural Networks (RNN) are powerful models that offer a compact, shared parametrization of a series of conditional distributions. RNNs have been shown to excel at hard sequence problems ranging from handwriting generation \cite{DBLP:journals/corr/Graves13}, to character prediction \cite{sutskever2011generating} and to machine translation \cite{kalchbrenner13emnlp}. A two-dimensional RNN has produced very promising results in modeling grayscale images and textures \cite{theis2015generative}. 

\begin{figure}[h]

\hfill
\begin{subfigure}{.14\textwidth}
	\includegraphics[width=\textwidth]{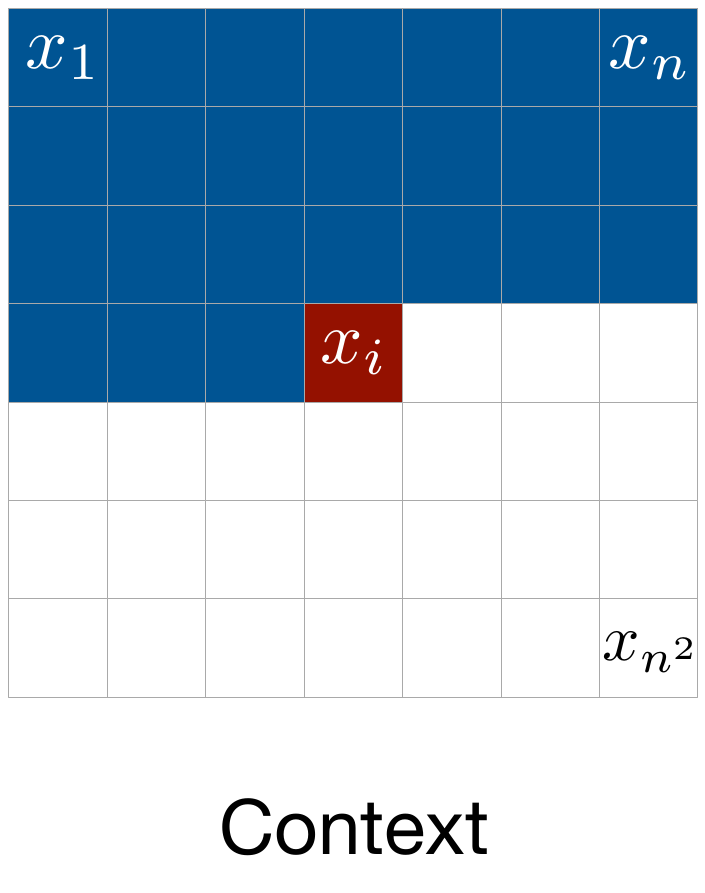}
\end{subfigure}
\hfill
\begin{subfigure}{.14\textwidth}
	\includegraphics[width=\textwidth]{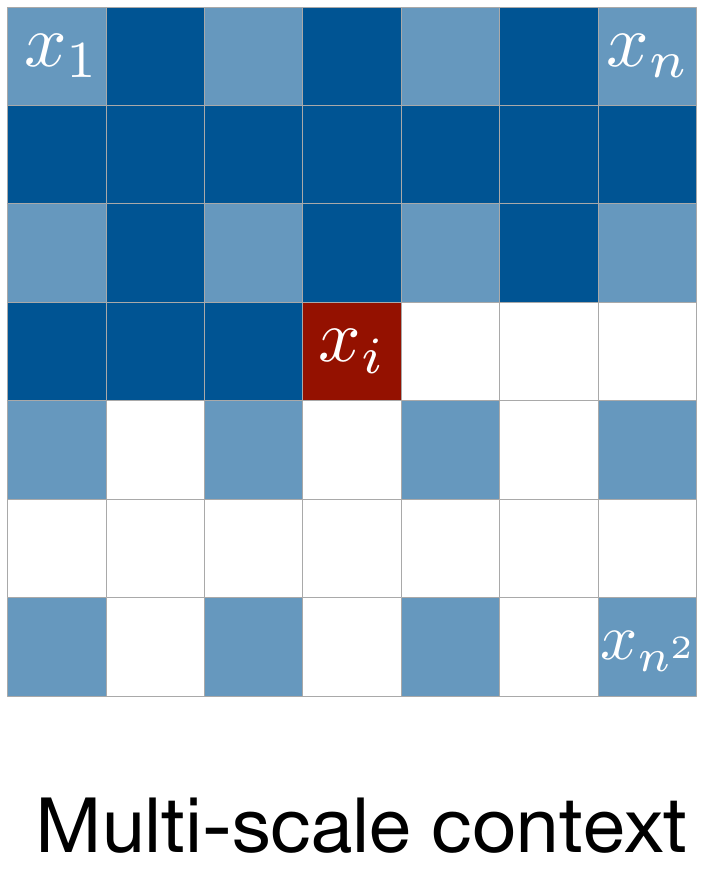}
\end{subfigure}
\hfill
\begin{subfigure}{.16\textwidth}
	\includegraphics[width=\textwidth]{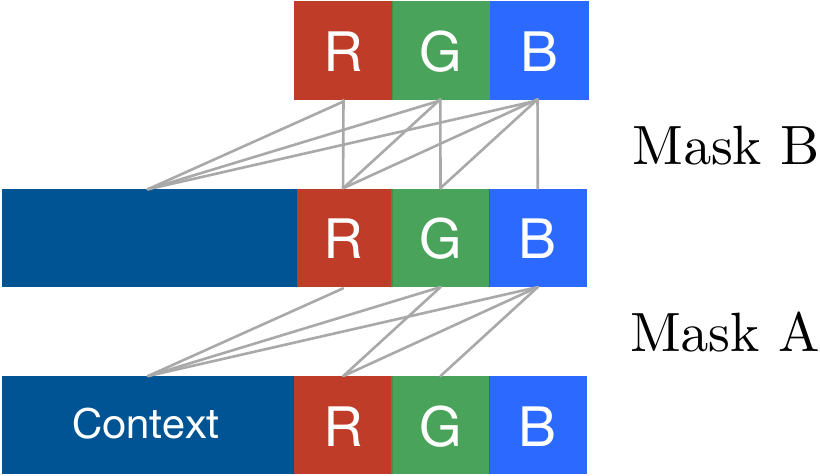}
\end{subfigure}
\hfill
\caption{\textbf{Left}: To generate pixel $x_i$ one conditions on all the previously generated pixels left and above of $x_i$. \textbf{Center}: To generate a pixel in the multi-scale case we can also condition on the subsampled image pixels (in light blue). \textbf{Right}: Diagram of the connectivity inside a masked convolution. In the first layer, each of the RGB channels is connected to previous channels and to the context, but is not connected to itself. In subsequent layers, the channels are also connected to themselves.}
\vspace{-0.2cm}
\label{depen}

\end{figure}

In this paper we advance two-dimensional RNNs and apply them to large-scale modeling of natural images.
The resulting \emph{PixelRNNs} are composed of up to twelve, fast two-dimensional Long Short-Term Memory (LSTM) layers. These layers use LSTM units in their state \cite{hochreiter1997long, graves2009offline}  and adopt a convolution to compute at once all the states along one of the spatial dimensions of the data. We design two types of these layers. The first type is the \emph{Row LSTM} layer where the convolution is applied along each row; a similar technique is described in \cite{NIPS2015_5642}. The second type is the \emph{Diagonal BiLSTM} layer where the convolution is applied in a novel fashion along the diagonals of the image. The networks also incorporate \emph{residual connections} \cite{DBLP:journals/corr/HeZRS15} around LSTM layers; we observe that this helps with training of the PixelRNN for up to twelve layers of depth. 

We also consider a second, simplified architecture which shares the same core components as the PixelRNN. 
We observe that Convolutional Neural Networks (CNN) can also be used as sequence model with a fixed dependency range, by using \emph{Masked} convolutions. The \emph{PixelCNN} architecture is a fully convolutional network of fifteen layers that preserves the spatial resolution of its input throughout the layers and outputs a conditional distribution at each location.
 
Both PixelRNN and PixelCNN capture the full generality of pixel inter-dependencies without introducing independence assumptions as in e.g., latent variable models. The dependencies are also maintained between the RGB color values within each individual pixel. 
Furthermore, in contrast to previous approaches that model the pixels as continuous values (e.g., \citet{theis2015generative, gregor2013deep}), we model the pixels as \emph{discrete} values using a multinomial distribution implemented with a simple softmax layer. We observe that this approach gives both representational and training advantages for our models.

The contributions of the paper are as follows. In Section \ref{sect:pixelrnn} we design two types of PixelRNNs corresponding to the two types of LSTM layers; we describe the purely convolutional PixelCNN that is our fastest architecture; and we design a \emph{Multi-Scale} version of the PixelRNN.
In Section \ref{sect:experiments} we show the relative benefits of using the discrete softmax distribution in our models and of adopting residual connections for the LSTM layers. Next we test the models on MNIST and on CIFAR-10 and show that they obtain log-likelihood scores that are considerably better than previous results. We also provide results for the large-scale ImageNet dataset resized to both $32 \times 32$ and $64 \times 64$ pixels; to our knowledge likelihood values from generative models have not previously been reported on this dataset. Finally, we give a qualitative evaluation of the samples generated from the PixelRNNs. 

\section{Model}
\label{framework}

Our aim is to estimate a distribution over natural images that can be used to tractably compute the likelihood of images and to generate new ones. The network scans the image one row at a time and one pixel at a time within each row.  For each pixel it predicts the conditional distribution over the possible pixel values given the scanned context. Figure \ref{depen} illustrates this process. The joint distribution over the image pixels is factorized into a product of conditional distributions. The parameters used in the predictions are shared across all pixel positions in the image. 

To capture the generation process, \citet{theis2015generative} propose to use a two-dimensional LSTM network \cite{graves2009offline} that starts at the top left pixel and proceeds towards the bottom right pixel. The advantage of the LSTM network is that it effectively handles long-range dependencies that are central to object and scene understanding. The two-dimensional structure ensures that the signals are well propagated both in the left-to-right and top-to-bottom directions. 

In this section we first focus on the form of the distribution,
whereas the next section will be devoted to describing the architectural innovations inside PixelRNN.

\subsection{Generating an Image Pixel by Pixel}

The goal is to assign a probability $p(\vec{x})$ to each image $\vec{x}$ formed of $n \times n$ pixels. We can write the image $\vec{x}$ as a one-dimensional sequence $x_1,...,x_{n^2}$ where pixels are taken from the image row by row. To estimate the joint distribution $p(\vec{x})$ we write it as the product of the conditional distributions over the pixels:
\vspace{-0.3cm}
\begin{equation}
p(\vec{x}) = \prod_{i=1}^{n^2} p(x_i | x_1,...,x_{i-1})
\end{equation}
The value $p(x_i | x_1,...,x_{i-1})$ is the probability of the $i$-th pixel $x_i$ given all the previous pixels $x_1,...,x_{i-1}$. The generation proceeds row by row and pixel by pixel. Figure \ref{depen} (Left) illustrates the conditioning scheme.

Each pixel $x_i$ is in turn jointly determined by three values, one for each of the color channels Red, Green and Blue (RGB). We rewrite the distribution $p(x_i|\vec{x}_{<i})$ as the following product:
\begin{equation}
p(x_{i,R}|\vec{x}_{<i})p(x_{i,G}|\vec{x}_{<i},x_{i,R})p(x_{i,B}|\vec{x}_{<i},x_{i,R},x_{i,G})
\label{eq:color_conditioning}
\end{equation} 
Each of the colors is thus conditioned on the other channels as well as on all the previously generated pixels. 

Note that during {training} and evaluation the distributions over the pixel values are computed \emph{in parallel}, while the generation of an image is \emph{sequential}. 

\subsection{Pixels as Discrete Variables}

Previous approaches use a continuous distribution for the values of the pixels in the image (e.g. \citet{theis2015generative, uria2013deep}). By contrast we model $p(\vec{x})$ as a discrete distribution, with every conditional distribution in Equation \ref{eq:color_conditioning} being a multinomial that is modeled with a softmax layer. Each channel variable $x_{i,*}$ simply takes one of 256 distinct values. The discrete distribution is representationally simple and has the advantage of being arbitrarily multimodal without prior on the shape (see Fig.~\ref{fig:softmax_activations}). Experimentally we also find the discrete distribution to be easy to learn and to produce better performance compared to a continuous distribution (Section \ref{sect:experiments}).

\section{Pixel Recurrent Neural Networks}
\label{sect:pixelrnn}

In this section we describe the architectural components that compose the {PixelRNN}. In Sections \ref{sect:row_lstm} and \ref{sect:diag_lstm}, we describe the two types of LSTM layers that use convolutions to compute at once the states along one of the spatial dimensions. 
In Section \ref{sect:residual} we describe how to incorporate \emph{residual} connections to improve the training of a PixelRNN with many LSTM layers.  In Section \ref{sect:masked} we describe the softmax layer that computes the {discrete} joint distribution of the colors and the masking technique that ensures the proper conditioning scheme. In Section \ref{sect:pixelcnn} we describe the PixelCNN architecture. Finally in Section \ref{sect:multiscale} we describe the multi-scale architecture.

\begin{figure}
\centering
\includegraphics[width=0.46\textwidth]{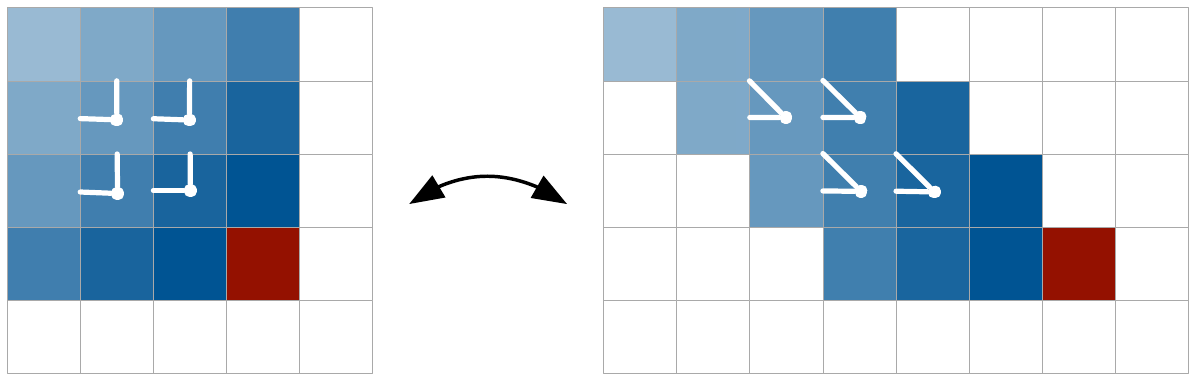}
\vspace{-0.3cm}
\caption{In the Diagonal BiLSTM, to allow for parallelization along the diagonals, the input map is skewed by offseting each row by one position with respect to the previous row. When the spatial layer is computed left to right and column by column, the output map is shifted back into the original size. The convolution uses a kernel of size $2 \times 1$. }
\label{fig:bilstm}
\vspace{-0.3cm}
\end{figure}

\subsection{Row LSTM}
\label{sect:row_lstm}

\begin{figure}[ht]
\includegraphics[width=0.48\textwidth]{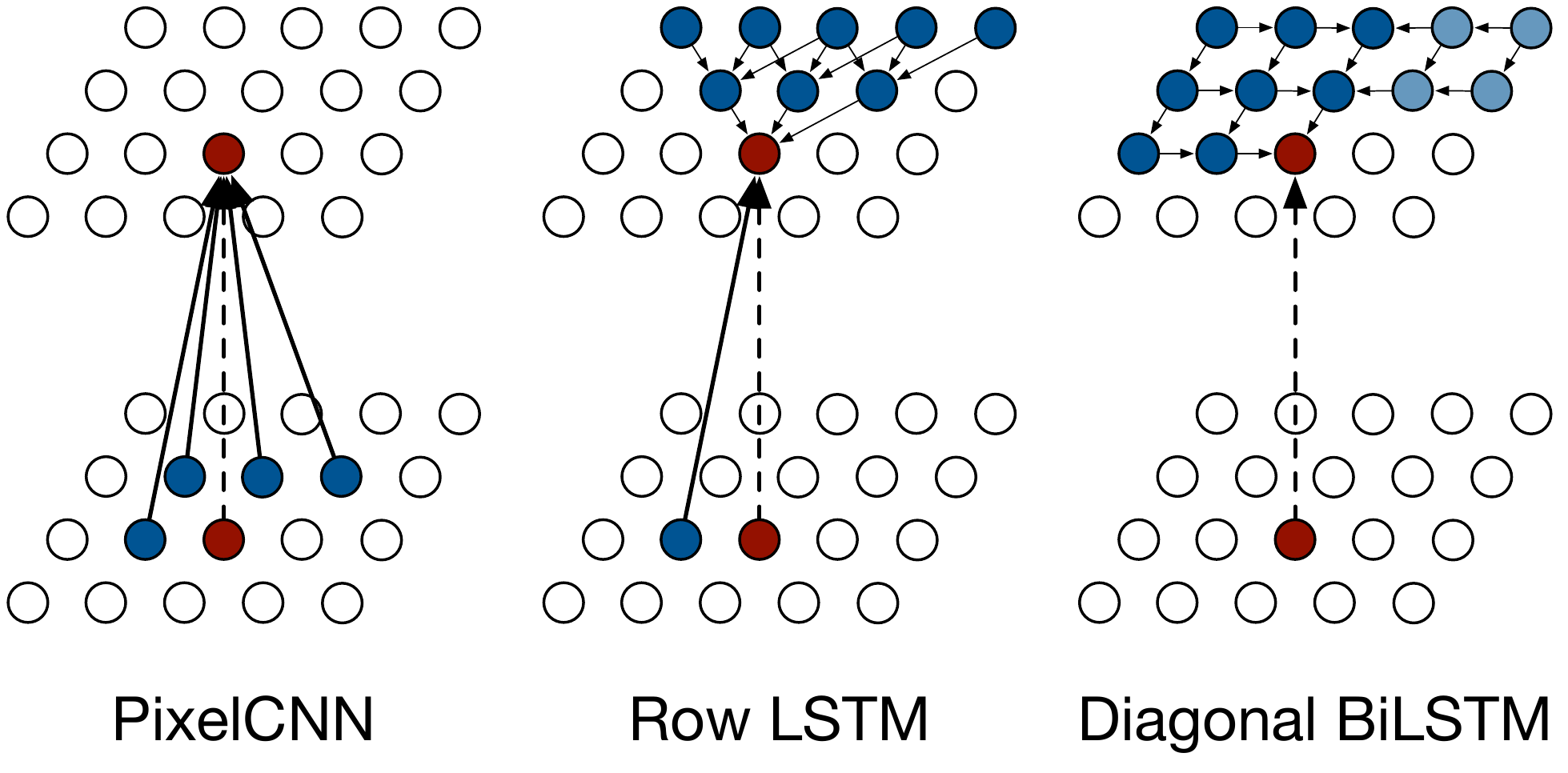}
\vspace{-0.7cm}
\caption{Visualization of the input-to-state and state-to-state mappings for the three proposed architectures.}
\label{mappings}
\vspace{-0.2cm}
\end{figure}

The Row LSTM is a unidirectional layer that processes the image row by row from top to bottom computing features for a whole row at once; the computation is performed with a one-dimensional convolution. For a pixel $x_i$ the layer captures a roughly triangular context above the pixel as shown in Figure \ref{mappings} (center). The kernel of the one-dimensional convolution has size $k \times 1$ where $k \geq 3$; the larger the value of $k$ the broader the context that is captured. The weight sharing in the convolution ensures translation invariance of the computed features along each row.

The computation proceeds as follows. An LSTM layer has an {input-to-state} component and a recurrent {state-to-state} component that together determine the four gates inside the LSTM core. To enhance parallelization in the Row LSTM the input-to-state component is first computed for the entire two-dimensional input map; for this a $k \times 1$ convolution is used to follow the row-wise orientation of the LSTM itself. The convolution is \emph{masked} to include only the valid context (see Section \ref{sect:masked}) and produces a tensor of size $4h \times n \times n$, representing the four gate vectors for each position in the input map, where $h$ is the number of output feature maps.

 To compute one step of the state-to-state component of the LSTM layer, one is given the previous hidden and cell states $\vec{h}_{i-1}$ and $\vec{c}_{i-1}$, each of size $h \times n \times 1$. The new hidden and cell states $\vec{h}_i$, $\vec{c}_i$ are obtained as follows:

 \vspace{-0.2cm}
 \begin{equation}
 \begin{split}
 [\vec{o}_i, \vec{f}_i, \vec{i}_i, \vec{g}_i ] & = \sigma (\vec{K}^{ss} \circledast \vec{h}_{i-1} + \vec{K}^{is} \circledast \vec{x}_{i}) \\
\vec{c}_i & = \vec{f}_i \odot \vec{c}_{i-1} + \vec{i}_i \odot \vec{g}_i \\
\vec{h}_i & = \vec{o}_i \odot \tanh(\vec{c}_i)\\
\end{split}
\label{eq:lstm}
 \vspace{-0.3cm}
 \end{equation}

where $\vec{x}_i$ of size $h \times n \times 1$ is row $i$ of the input map, and $\circledast$ represents the convolution operation and $\odot$ the element-wise multiplication. The weights $\vec{K}^{ss}$ and $\vec{K}^{is}$ are the kernel weights for the state-to-state and the input-to-state components, where the latter is precomputed as described above. In the case of the output, forget and input gates $\vec{o}_i$, $\vec{f}_i$ and $\vec{i}_i$, the activation $\sigma$ is the logistic sigmoid function, whereas for the content gate $\vec{g}_i$, $\sigma$ is the $\tanh$ function. Each step computes at once the new state for an entire row of the input map. Because the Row LSTM has a triangular receptive field (Figure \ref{mappings}), it is unable to capture the entire available context.

\subsection{Diagonal BiLSTM}
\label{sect:diag_lstm}

The Diagonal BiLSTM is designed to both parallelize the computation and to capture the entire available context for any image size. Each of the two directions of the layer scans the image in a diagonal fashion starting from a corner at the top and reaching the opposite corner at the bottom. Each step in the computation computes at once the LSTM state along a diagonal in the image. Figure \ref{mappings} (right) illustrates the computation and the resulting receptive field.

The diagonal computation proceeds as follows. We first skew the input map into a space that makes it easy to apply convolutions along diagonals. The {skewing} operation offsets each row of the input map by one position with respect to the previous row, as illustrated in Figure \ref{fig:bilstm}; this results in a map of size $n \times (2n-1)$. At this point we can compute the input-to-state and state-to-state components of the Diagonal BiLSTM. For each of the two directions, the input-to-state component is simply a $1 \times 1$ convolution $K^{is}$ that contributes to the four gates in the LSTM core; the operation generates a $4 h \times n \times n$ tensor. The state-to-state recurrent component is then computed with a \emph{column-wise} convolution $K^{ss}$ that has a kernel of size $2 \times 1$. The step takes the previous hidden and cell states, combines the contribution of the input-to-state component and produces the next hidden and cell states, as defined in Equation \ref{eq:lstm}. The output feature map is then skewed back into an $n \times n$ map by removing the offset positions. This computation is repeated for each of the two directions. Given the two output maps, to prevent the layer from seeing future pixels, the \emph{right} output map is then shifted down by one row and added to the \emph{left} output map.
 
Besides reaching the full dependency field, the Diagonal BiLSTM has the additional advantage that it uses a convolutional kernel of size $2\times1$ that processes a minimal amount of information at each step yielding a highly non-linear computation. Kernel sizes larger than $2\times1$ are not particularly useful as they do not broaden the already global receptive field of the Diagonal BiLSTM. 

\subsection{Residual Connections}
\label{sect:residual}

\begin{figure}[ht]
\centering
\begin{subfigure}{.4\textwidth}
  \centering
  \includegraphics[width=\linewidth]{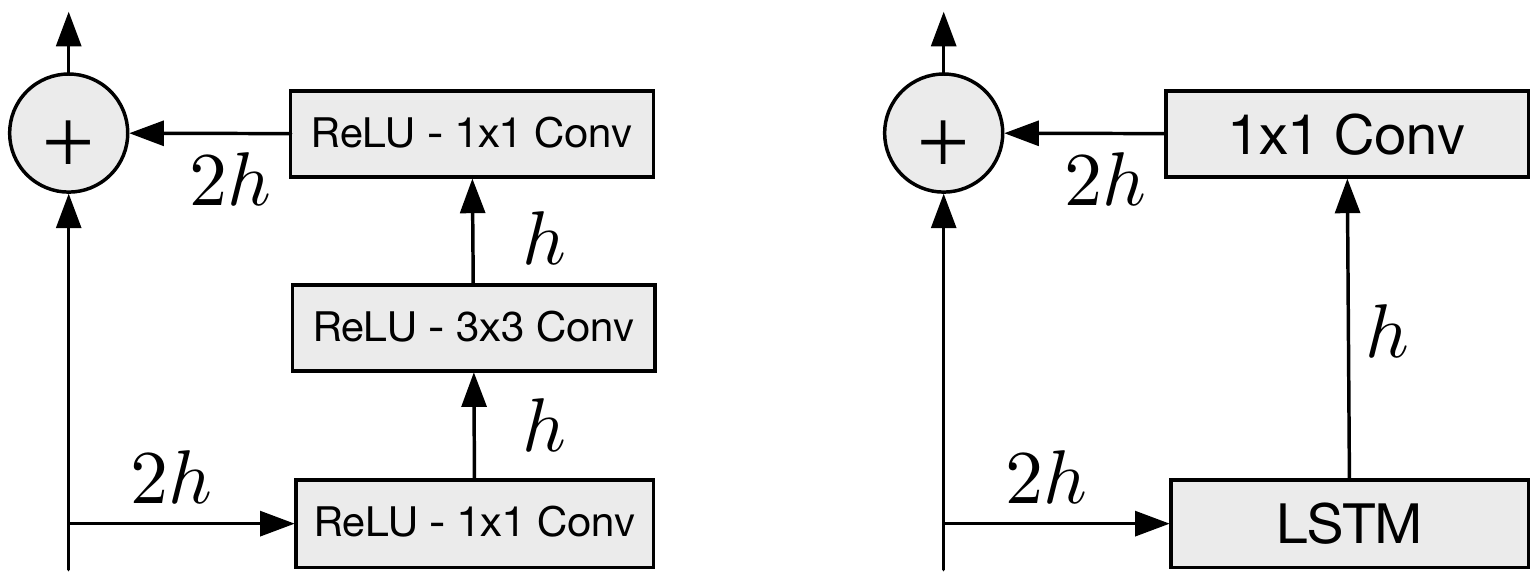}
\end{subfigure}%
\caption{Residual blocks for a PixelCNN (left) and PixelRNNs.}
\label{fig:residual_blocks}
\vspace{-0.2cm}
\end{figure}

We train PixelRNNs of up to twelve layers of depth. As a means to both increase convergence speed and propagate signals more directly through the network, we deploy \emph{residual connections} \cite{DBLP:journals/corr/HeZRS15} from one LSTM layer to the next. Figure \ref{fig:residual_blocks} shows a diagram of the residual blocks. The input map to the PixelRNN LSTM layer has $2h$ features. The input-to-state component reduces the number of features by producing $h$ features per gate. After applying the recurrent layer, the output map is upsampled back to $2h$ features per position via a $1\times1 $ convolution and the input map is added to the output map. This method is related to previous approaches that use gating along the depth of the recurrent network \cite{DBLP:journals/corr/KalchbrennerDG15,zhang2016highway}, but has the advantage of not requiring additional gates. Apart from residual connections, one can also use learnable skip connections from each layer to the output. In the experiments we evaluate the relative effectiveness of residual and layer-to-output skip connections. 
	
\subsection{Masked Convolution}
\label{sect:masked}

The $h$ features for each input position at every layer in the network are split into three parts, each corresponding to one of the RGB channels. When predicting the R channel for the current pixel $x_i$, only the generated pixels left and above of $x_i$ can be used as context. When predicting the G channel, the value of the R channel can also be used as context in addition to the previously generated pixels. Likewise, for the B channel, the values of both the R and G channels can be used. To restrict connections in the network to these dependencies, we apply a \emph{mask} to the input-to-state convolutions and to other purely convolutional layers in a PixelRNN. 

We use two types of masks that we indicate with \emph{mask A} and \emph{mask B}, as shown in Figure \ref{depen} (Right). Mask A is applied only to the first convolutional layer in a PixelRNN and restricts the connections to those neighboring pixels and to those colors in the current pixels that have already been predicted. On the other hand, mask B is applied to all the subsequent input-to-state convolutional transitions and relaxes the restrictions of mask A by also allowing the connection from a color to itself. The masks can be easily implemented by zeroing out the corresponding weights in the input-to-state convolutions after each update. Similar masks have also been used in variational autoencoders \cite{gregor2013deep, germain2015made}.

\begin{table}[t]
\small
	\begin{center}
	\begin{tabular}{l|l|l}
		\toprule
		\multicolumn{1}{c|}{\textbf{ PixelCNN} }& \multicolumn{1}{c|}{\textbf{ Row LSTM} } & \multicolumn{1}{|c}{\textbf{ Diagonal BiLSTM} }  \\ \midrule
		\multicolumn{3}{c}{ $7 \times 7$ conv mask A} \\ \midrule 
		\multicolumn{3}{c}{ \textbf{Multiple residual blocks:} (see fig \ref{fig:residual_blocks})} \\ 
		\multicolumn{3}{c}{ } \\ 
	    \multirow{3}{*}{}Conv & Row LSTM & Diagonal BiLSTM \\ 
	    				 $3 \times 3$  mask B & i-s: $3 \times 1$ mask B & i-s: $1\times1$ mask B \\
	    				 & s-s: $3 \times 1$ no mask & s-s: $1\times2$ no mask \\ \midrule
	    \multicolumn{3}{c}{ ReLU followed by $1 \times 1$ conv, mask B (2 layers)} \\ \midrule
	    \multicolumn{3}{c}{ 256-way Softmax for each RGB color (Natural images)}\\
	    \multicolumn{3}{c}{ or Sigmoid (MNIST)} \\ 
	    \bottomrule
	\end{tabular}
	\end{center}
\vspace{-0.2cm}
\caption{Details of the architectures. In the LSTM architectures i-s and s-s stand for input-state and state-state convolutions.}
\vspace{-0.3cm}
\label{table:architectures}
\end{table}

\subsection{PixelCNN}
\label{sect:pixelcnn}

The Row and Diagonal LSTM layers have a potentially unbounded dependency range within their receptive field. This comes with a computational cost as each state needs to be computed sequentially. One simple workaround is to make the receptive field large, but not unbounded. We can use standard convolutional layers to capture a bounded receptive field and compute features for all pixel positions at once. The PixelCNN uses multiple convolutional layers that preserve the spatial resolution; pooling layers are not used. Masks are adopted in the convolutions to avoid seeing the future context; masks have previously also been used in non-convolutional models such as MADE \cite{germain2015made}. Note that the advantage of parallelization of the PixelCNN over the PixelRNN is only available during training or during evaluating of test images. The image generation process is sequential for both kinds of networks, as each sampled pixel needs to be given as input back into the network.

\subsection{Multi-Scale PixelRNN}
\label{sect:multiscale}

The Multi-Scale PixelRNN is composed of an \emph{unconditional} PixelRNN and one or more \emph{conditional} PixelRNNs. The unconditional network first generates in the standard way a smaller $s \times s$ image that is \emph{subsampled} from the original image. The conditional network then takes the $s \times s$ image as an additional input and generates a larger $n\times n$ image, as shown in Figure \ref{depen} (Middle).

The conditional network is similar to a standard PixelRNN, but each of its layers is biased with an upsampled version of the small $s \times s$ image. The upsampling and biasing processes are defined as follows. In the upsampling process, one uses a convolutional network with deconvolutional layers to construct an enlarged feature map of size $c \times n \times n$, where $c$ is the number of features in the output map of the upsampling network. Then, in the biasing process, for each layer in the conditional PixelRNN, one simply maps the $c \times n \times n$ conditioning map into a $4h \times n \times n$ map that is added to the input-to-state map of the corresponding layer; this is performed using a $1 \times 1$ unmasked convolution. The larger $n \times n$ image is then generated as usual.

\section{Specifications of Models}

In this section we give the specifications of the PixelRNNs used in the experiments. We have four types of networks: the PixelRNN based on Row LSTM, the one based on Diagonal BiLSTM, the fully convolutional one and the Multi-Scale one. 

Table \ref{table:architectures} specifies each layer in the single-scale networks. The first layer is a $7\times7$ convolution that uses the mask of type A. The two types of LSTM networks then use a variable number of recurrent layers. The input-to-state convolution in this layer uses a mask of type B, whereas the state-to-state convolution is not masked. The PixelCNN uses convolutions of size $3\times3$ with a mask of type B. The top feature map is then passed through a couple of layers consisting of a Rectified Linear Unit (ReLU) and a $1\times1$ convolution. For the CIFAR-10 and ImageNet experiments, these layers have 1024 feature maps; for the MNIST experiment, the layers have 32 feature maps. Residual and layer-to-output connections are used across the layers of all three networks.

The networks used in the experiments have the following hyperparameters. For MNIST we use a Diagonal BiLSTM with 7 layers and a value of $h=16$  (Section \ref{sect:residual} and Figure \ref{fig:residual_blocks} right). For CIFAR-10 the Row and Diagonal BiLSTMs have 12 layers and a number of $h=128$ units. The PixelCNN has 15 layers and $h=128$. For $32\times32$ ImageNet  we adopt a 12 layer Row LSTM with $h=384$ units and for $64\times64$ ImageNet we use a 4 layer Row LSTM with $h=512$ units; the latter model does not use residual connections. 


\section{Experiments}
\label{sect:experiments}

In this section we describe our experiments and results.  We begin by describing the way we evaluate and compare our results. In Section \ref{sect:training_details} we give details about the training. Then we give results on the relative effectiveness of architectural components and our best results on the MNIST, CIFAR-10 and ImageNet datasets.

\subsection{Evaluation}

All our models are trained and evaluated on the log-likelihood loss function coming from a discrete distribution.
Although natural image data is usually modeled with \emph{continuous} distributions using density functions, we can compare our results with previous art in the following way. In the literature it is currently best practice to add real-valued noise to the pixel values to dequantize the data when using density functions \cite{uria2013rnade}. When uniform noise is added (with values in the interval [0, 1]), then the log-likelihoods of continuous and discrete models are directly comparable \cite{theis2015note}. In our case, we can use the values from the discrete distribution as a piecewise-uniform continuous function that has a constant value for every interval $[i, i+1], i = 1, 2, \dots 256$. This corresponding distribution will have the same log-likelihood (on data with added noise) as the original discrete distribution (on discrete data). 

For MNIST we report the negative log-likelihood in \emph{nats} as it is common practice in literature. For CIFAR-10 and ImageNet we report negative log-likelihoods in \emph{bits} per dimension. The total discrete log-likelihood is normalized by the dimensionality of the images (e.g., $32\times 32\times 3=3072$ for CIFAR-10). These numbers are interpretable as the number of bits that a compression scheme based on this model would need to compress every RGB color value \mbox{\citep{van2014student, theis2015note}}; in practice there is also a small overhead due to arithmetic coding.

\subsection{Training Details}
\label{sect:training_details}

Our models are trained on GPUs using the Torch toolbox. From the different parameter update rules tried, RMSProp gives best convergence performance and is used for all experiments. The learning rate schedules were manually set for every dataset to the highest values that allowed fast convergence. The batch sizes also vary for different datasets. For smaller datasets such as MNIST and CIFAR-10 we use smaller batch sizes of 16 images as this seems to regularize the models. For ImageNet we use as large a batch size as allowed by the GPU memory; this corresponds to 64 images/batch for $32\times32$ ImageNet, and 32 images/batch for $64\times64$ ImageNet. Apart from scaling and centering the images at the input of the network, we don't use any other preprocessing or augmentation. For the multinomial loss function we use the raw pixel color values as categories. For all the PixelRNN models, we learn the initial recurrent state of the network.

\subsection{Discrete Softmax Distribution}
Apart from being intuitive and easy to implement, we find that using a softmax on discrete pixel values instead of a mixture density approach on continuous pixel values gives better results. For the Row LSTM model with a softmax output distribution we obtain 3.06 bits/dim on the CIFAR-10 validation set. For the same model with a Mixture of Conditional Gaussian Scale Mixtures (MCGSM) \cite{theis2015generative} we obtain 3.22 bits/dim. 

\begin{figure}[h]
  \centering
  \includegraphics[width=0.48\textwidth]{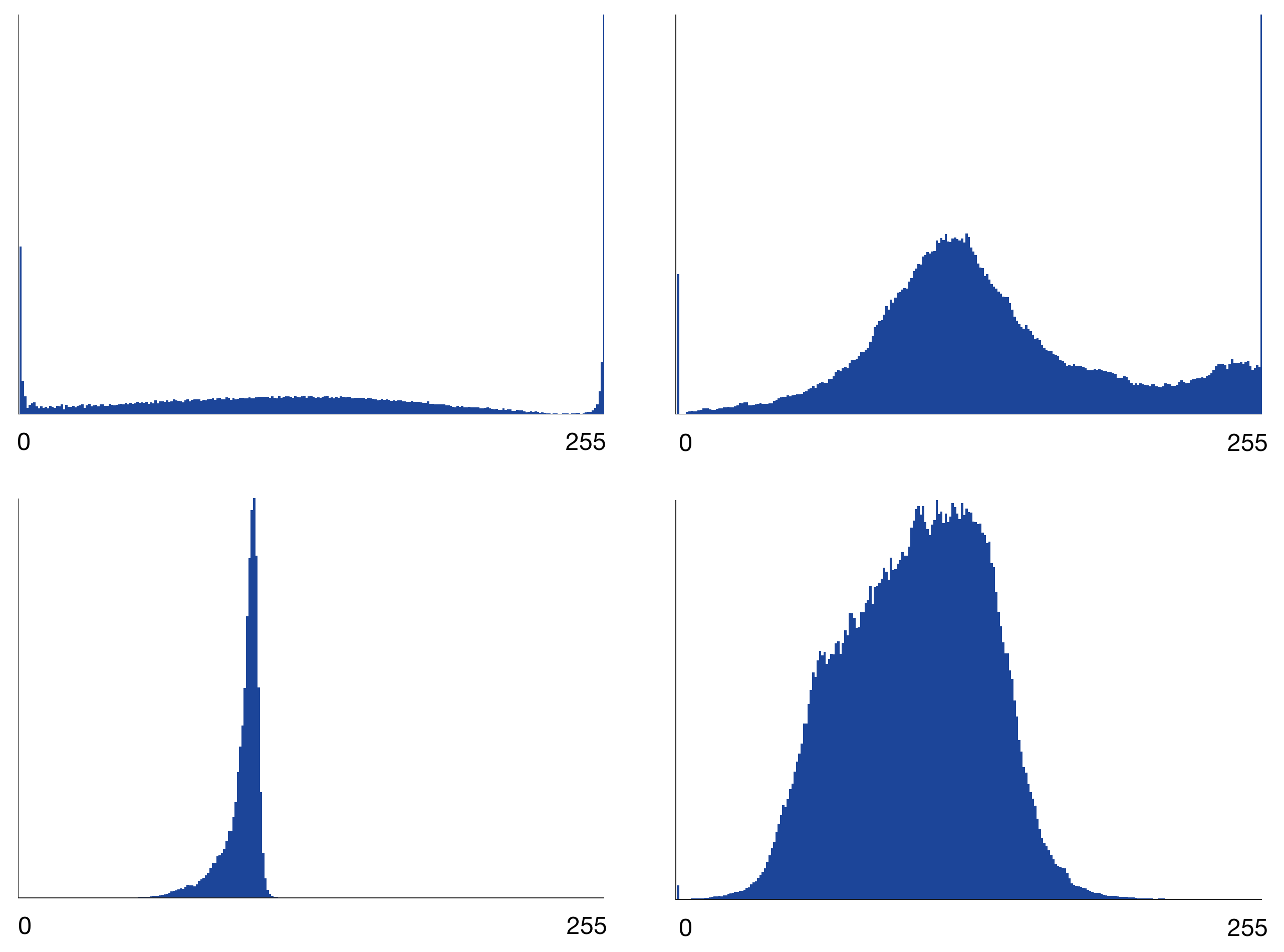}
  \vspace{-0.5cm}
  \caption{Example softmax activations from the model. The top left shows the distribution of the first pixel red value (first value to sample).}
  \label{fig:softmax_activations}
\end{figure}

In Figure \ref{fig:softmax_activations} we show a few softmax activations from the model. Although we don't embed prior information about the meaning or relations of the 256 color categories, e.g. that pixel values 51 and 52 are neighbors, the distributions predicted by the model are meaningful and can be multimodal, skewed, peaked or long tailed. Also note that values 0 and 255 often get a much higher probability as they are more frequent. Another advantage of the discrete distribution is that we do not worry about parts of the distribution mass lying outside the interval [0, 255], which is something that typically happens with continuous distributions.

\begin{figure*}[ht]
\begin{subfigure}{.5\textwidth}
  \centering
  \includegraphics[trim={0 0 0 0},clip, width=0.95\linewidth]{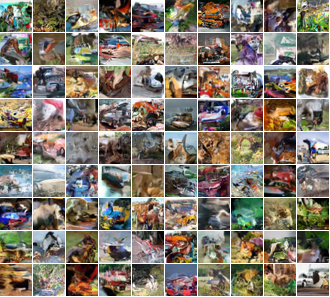}
\end{subfigure}%
\hfill
\begin{subfigure}{.5\textwidth}
  \centering
  \includegraphics[trim={0 0 0 0},clip, width=0.95\linewidth]{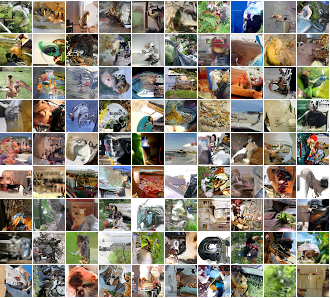}
\end{subfigure}%
\hfill
\caption{Samples from models trained on CIFAR-10 (left) and ImageNet 32x32 (right) images. In general we can see that the models capture local spatial dependencies relatively well. The ImageNet model seems to be better at capturing more global structures than the CIFAR-10 model. The ImageNet model was larger and trained on much more data, which explains the qualitative difference in samples.}
\label{fig:samples_32}
\end{figure*}

\subsection{Residual Connections}
Another core component of the networks is residual connections. In Table \ref{table:effect_skipconnections} we show the results of having residual connections, having standard skip connections or having both, in the 12-layer CIFAR-10 Row LSTM model. We see that using residual connections is as effective as using skip connections; using both is also effective and preserves the advantage. 

\begin{table}[h]
	\begin{center}
	\begin{tabular}{ccc}
		\toprule
		 & \textbf{No skip} & \textbf{Skip} \\ 
		\midrule
		\textbf{No residual}: & 3.22 & 3.09 \\ 
	    \textbf{Residual}: & 3.07 & 3.06 \\ 
	    \bottomrule
	\end{tabular}
	\end{center}
\vspace{-0.3cm}
\caption{Effect of residual and skip connections in the Row LSTM network evaluated on the Cifar-10 validation set in bits/dim.}
\label{table:effect_skipconnections}
\end{table}

When using both the residual and skip connections, we see in Table \ref{table:effect_of_layers} that performance of the Row LSTM improves with increased depth. This holds for up to the 12 LSTM layers that we tried.

\begin{table}[h]
\centering
	\begin{tabular}{ccccccc}
		\toprule
		\textbf{\# layers}: & 1 & 2 & 3 & 6 & 9 & 12 \\ 
	    \midrule
	    \textbf{NLL}: & 3.30 & 3.20 & 3.17 & 3.09 & 3.08 & 3.06 \\
	    \bottomrule
	\end{tabular}
\caption{Effect of the number of layers on the negative log likelihood evaluated on the CIFAR-10 validation set (bits/dim).}
\label{table:effect_of_layers}
\end{table}

\subsection{MNIST}

Although the goal of our work was to model natural images on a large scale, we also tried our model on the binary version \cite{salakhutdinov2008quantitative} of MNIST \cite{lecun1998gradient} as it is a good sanity check and there is a lot of previous art on this dataset to compare with. In Table \ref{table:mnist} we report the performance of the Diagonal BiLSTM model and that of previous published results. To our knowledge this is the best reported result on MNIST so far. 

\begin{table}[!h]
	\begin{center}
	\begin{tabular}{lcc}
		\toprule
		\textbf{Model} & \textbf{NLL Test}  \\ 
		\midrule
		DBM 2hl [1]: & $\approx$ 84.62 \\ 
		DBN 2hl [2]: & $\approx$ 84.55 \\ 
		NADE [3]: & 88.33 \\ 
		EoNADE 2hl (128 orderings) [3]: & 85.10 \\ 
		EoNADE-5 2hl (128 orderings) [4]: \quad \quad \quad & 84.68 \\ 
		DLGM [5]: & $\approx$ 86.60 \\ 
		DLGM 8 leapfrog steps [6]: & $\approx$ 85.51 \\ 
		DARN 1hl [7]: & $\approx$ 84.13 \\ 
		MADE 2hl (32 masks) [8]: & 86.64 \\
		DRAW [9]: & $\leq$ 80.97 \\ 
		\midrule
		PixelCNN: & 81.30 \\ 
		Row LSTM: & 80.54 \\ 
		Diagonal BiLSTM (1 layer, $h=32$): & \textbf{80.75} \\
		Diagonal BiLSTM (7 layers, $h=16$): & \textbf{79.20} \\ 
	    \bottomrule
	\end{tabular}
	\end{center}
\vspace{-0.2cm}
\caption{Test set performance of different models on MNIST in \emph{nats} (negative log-likelihood). Prior results taken from [1] \cite{salakhutdinov2009deep}, [2] \cite{murray2009evaluating}, [3] \cite{uria2013deep}, [4] \cite{raiko2014iterative}, [5] \cite{rezende2014stochastic}, [6] \cite{salimans2014markov}, [7] \cite{gregor2013deep}, [8] \cite{germain2015made}, [9] \cite{gregor2015draw}.}
\label{table:mnist}
\end{table}

\begin{table}[!h]
\centering
	\begin{tabular}{lcc}
		\toprule
		\textbf{Model} & \textbf{NLL Test (Train)}  \\ 
		\midrule
		Uniform Distribution: & 8.00 \\ 
		Multivariate Gaussian: & 4.70 \\ 
		NICE [1]: & 4.48 \\ 
		Deep Diffusion [2]: & 4.20 \\ 
		Deep GMMs [3]: & 4.00 \\
		RIDE [4]: & 3.47 \\ 
		\midrule
		PixelCNN: & 3.14 (3.08) \\ 
		Row LSTM: & 3.07 (3.00) \\ 
		Diagonal BiLSTM: \quad\quad\quad\quad\quad\quad\quad & \textbf{3.00} (2.93) \\ 
	    \bottomrule
	\end{tabular}
\caption{Test set performance of different models on CIFAR-10 in \emph{bits/dim}. For our models we give training performance in brackets. [1] \cite{dinh2014nice}, [2] \cite{deepdiffusion}, [3] \cite{van2014factoring}, [4] personal communication \cite{theis2015generative}.}
\label{table:cifar10}
\end{table}

\begin{table}[!h]
	\begin{center}
	\begin{tabular}{lcc}
		\toprule
		\textbf{Image size} & \textbf{NLL Validation (Train)}  \\ 
		\midrule
		32x32: & 3.86 (3.83) \\ 
		64x64: & 3.63 (3.57) \\ 
	    \bottomrule
	\end{tabular}
	\end{center}
\vspace{-0.2cm}
\caption{Negative log-likelihood performance on $32\times32$ and $64\times64$ ImageNet in \emph{bits/dim}.}
\label{table:imagenet}
\vspace{-0.4cm}
\end{table}

\begin{figure*}[ht]

\begin{subfigure}{.5\textwidth}
  \centering
  \includegraphics[trim={0 0 0 0},clip, width=0.95\textwidth]{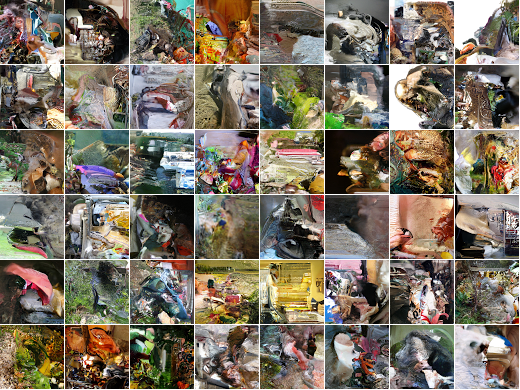}
\end{subfigure}%
\hfill
\begin{subfigure}{.5\textwidth}
  \centering
  \includegraphics[trim={0 0 0 0},clip, width=0.95\linewidth]{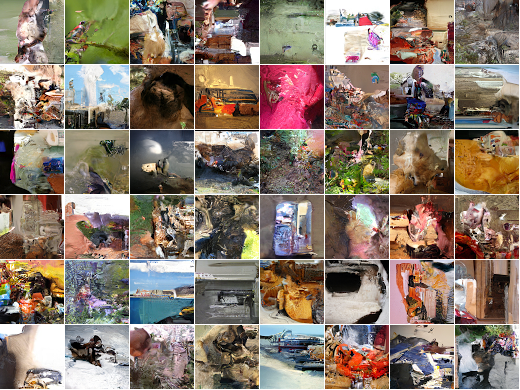}
\end{subfigure}%
\hfill
\caption{Samples from models trained on ImageNet 64x64 images. Left: normal model, right: multi-scale model. The single-scale model trained on 64x64 images is less able to capture global structure than the 32x32 model. The multi-scale model seems to resolve this problem. Although these models get similar performance in log-likelihood, the samples on the right do seem globally more coherent.}
\label{fig:samples_64}
\end{figure*}

\subsection{CIFAR-10}

Next we test our models on the CIFAR-10 dataset \cite{krizhevsky2009learning}. Table \ref{table:cifar10} lists the results of our models and that of previously published approaches. All our results were obtained without data augmentation. For the proposed networks, the Diagonal BiLSTM has the best performance, followed by the Row LSTM and the PixelCNN. This coincides with the size of the respective receptive fields: the Diagonal BiLSTM has a global view, the Row LSTM has a partially occluded view and the PixelCNN sees the fewest pixels in the context. This suggests that effectively capturing a large receptive field is important. 
Figure \ref{fig:samples_32} (left) shows CIFAR-10 samples generated from the Diagonal BiLSTM.

\begin{figure}[!ht]
\centering
\hspace{0.02cm} {occluded} \hfill completions \hfill{original} \,
\vspace{0.1cm}
\includegraphics[width=\linewidth]{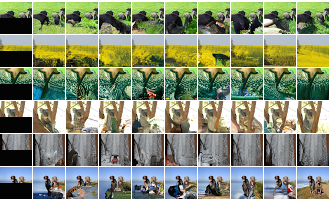}
\vspace{-0.5cm}
\caption{Image completions sampled from a model that was trained on 32x32 ImageNet images. Note that diversity of the completions is high, which can be attributed to the log-likelihood loss function used in this generative model, as it encourages models with high entropy. As these are sampled from the model, we can easily generate millions of different completions. It is also interesting to see that textures such as water, wood and shrubbery are also inputed relative well (see Figure \ref{fig:intro_completions}).}
\label{fig:completions}
\vspace{-0.2cm}
\end{figure}



\vspace{0.4cm}
\subsection{ImageNet}

Although to our knowledge the are no published results on the ILSVRC ImageNet dataset \cite{ILSVRC15} that we can compare our models with, we give our ImageNet log-likelihood performance in Table \ref{table:imagenet} (without data augmentation). On ImageNet the current PixelRNNs do not appear to overfit, as we saw that their validation performance improved with size and depth. The main constraint on model size are currently computation time and GPU memory.

Note that the ImageNet models are in general less compressible than the CIFAR-10 images. ImageNet has greater variety of images, and the CIFAR-10 images were most likely resized with a different algorithm than the one we used for ImageNet images. The ImageNet images are less blurry, which means neighboring pixels are less correlated to each other and thus less predictable. Because the downsampling method can influence the compression performance, we have made the used downsampled images available\footnote{http://image-net.org/small/download.php}.

Figure \ref{fig:samples_32} (right) shows $32\times32$ samples drawn from our model trained on ImageNet. Figure \ref{fig:samples_64} shows $64\times64$ samples from the same model with and without multi-scale conditioning. Finally, we also show image completions sampled from the model in Figure \ref{fig:completions}.


\section{Conclusion}

In this paper we significantly improve and build upon deep recurrent neural networks as generative models for natural images. We have described novel two-dimensional LSTM layers: the Row LSTM and the Diagonal BiLSTM, that scale more easily to larger datasets. The models were trained to model the raw RGB pixel values. We treated the pixel values as discrete random variables by using a softmax layer in the conditional distributions. We employed masked convolutions to allow PixelRNNs to model full dependencies between the color channels. We proposed and evaluated architectural improvements in these models resulting in PixelRNNs with up to 12 LSTM layers.

We have shown that the PixelRNNs significantly improve the state of the art on the MNIST and CIFAR-10 datasets. We also provide new benchmarks for generative image modeling on the ImageNet dataset. Based on the samples and completions drawn from the models we can conclude that the PixelRNNs are able to model both spatially local and long-range correlations and are able to produce images that are sharp and coherent.
Given that these models improve as we make them larger and that there is practically unlimited data available to train on, more computation and larger models are likely to further improve the results.



\section*{Acknowledgements}

The authors would like to thank Shakir Mohamed and Guillaume Desjardins for helpful input on this paper and Lucas Theis, Alex Graves, Karen Simonyan, Lasse Espeholt, Danilo Rezende, Karol Gregor and Ivo Danihelka for insightful discussions.

\bibliography{pixelRNN}

\begin{thebibliography}{33}
\providecommand{\natexlab}[1]{#1}
\providecommand{\url}[1]{\texttt{#1}}
\expandafter\ifx\csname urlstyle\endcsname\relax
  \providecommand{\doi}[1]{doi: #1}\else
  \providecommand{\doi}{doi: \begingroup \urlstyle{rm}\Url}\fi

\bibitem[Bengio \& Bengio(2000)Bengio and Bengio]{Bengio_Bengio_NIPS99}
Bengio, Yoshua and Bengio, Samy.
\newblock Modeling high-dimensional discrete data with multi-layer neural
  networks.
\newblock pp.\  400--406. MIT Press, 2000.

\bibitem[Dinh et~al.(2014)Dinh, Krueger, and Bengio]{dinh2014nice}
Dinh, Laurent, Krueger, David, and Bengio, Yoshua.
\newblock {NICE}: Non-linear independent components estimation.
\newblock \emph{arXiv preprint arXiv:1410.8516}, 2014.

\bibitem[Germain et~al.(2015)Germain, Gregor, Murray, and
  Larochelle]{germain2015made}
Germain, Mathieu, Gregor, Karol, Murray, Iain, and Larochelle, Hugo.
\newblock {MADE}: Masked autoencoder for distribution estimation.
\newblock \emph{arXiv preprint arXiv:1502.03509}, 2015.

\bibitem[Graves(2013)]{DBLP:journals/corr/Graves13}
Graves, Alex.
\newblock Generating sequences with recurrent neural networks.
\newblock \emph{arXiv preprint arXiv:1308.0850}, 2013.

\bibitem[Graves \& Schmidhuber(2009)Graves and Schmidhuber]{graves2009offline}
Graves, Alex and Schmidhuber, J{\"u}rgen.
\newblock Offline handwriting recognition with multidimensional recurrent
  neural networks.
\newblock In \emph{Advances in Neural Information Processing Systems}, 2009.

\bibitem[Gregor et~al.(2014)Gregor, Danihelka, Mnih, Blundell, and
  Wierstra]{gregor2013deep}
Gregor, Karol, Danihelka, Ivo, Mnih, Andriy, Blundell, Charles, and Wierstra,
  Daan.
\newblock Deep autoregressive networks.
\newblock In \emph{Proceedings of the 31st International Conference on Machine
  Learning}, 2014.

\bibitem[Gregor et~al.(2015)Gregor, Danihelka, Graves, and
  Wierstra]{gregor2015draw}
Gregor, Karol, Danihelka, Ivo, Graves, Alex, and Wierstra, Daan.
\newblock {DRAW}: A recurrent neural network for image generation.
\newblock \emph{Proceedings of the 32nd International Conference on Machine
  Learning}, 2015.

\bibitem[He et~al.(2015)He, Zhang, Ren, and Sun]{DBLP:journals/corr/HeZRS15}
He, Kaiming, Zhang, Xiangyu, Ren, Shaoqing, and Sun, Jian.
\newblock Deep residual learning for image recognition.
\newblock \emph{arXiv preprint arXiv:1512.03385}, 2015.

\bibitem[Hochreiter \& Schmidhuber(1997)Hochreiter and
  Schmidhuber]{hochreiter1997long}
Hochreiter, Sepp and Schmidhuber, J{\"u}rgen.
\newblock Long short-term memory.
\newblock \emph{Neural computation}, 1997.

\bibitem[Kalchbrenner \& Blunsom(2013)Kalchbrenner and
  Blunsom]{kalchbrenner13emnlp}
Kalchbrenner, Nal and Blunsom, Phil.
\newblock Recurrent continuous translation models.
\newblock In \emph{Proceedings of the 2013 Conference on Empirical Methods in
  Natural Language Processing}, 2013.

\bibitem[Kalchbrenner et~al.(2015)Kalchbrenner, Danihelka, and
  Graves]{DBLP:journals/corr/KalchbrennerDG15}
Kalchbrenner, Nal, Danihelka, Ivo, and Graves, Alex.
\newblock Grid long short-term memory.
\newblock \emph{arXiv preprint arXiv:1507.01526}, 2015.

\bibitem[Kingma \& Welling(2013)Kingma and
  Welling]{DBLP:journals/corr/KingmaW13}
Kingma, Diederik~P and Welling, Max.
\newblock Auto-encoding variational bayes.
\newblock \emph{arXiv preprint arXiv:1312.6114}, 2013.

\bibitem[Krizhevsky(2009)]{krizhevsky2009learning}
Krizhevsky, Alex.
\newblock Learning multiple layers of features from tiny images.
\newblock 2009.

\bibitem[Larochelle \& Murray(2011)Larochelle and Murray]{larochelle2011}
Larochelle, Hugo and Murray, Iain.
\newblock The neural autoregressive distribution estimator.
\newblock \emph{The Journal of Machine Learning Research}, 2011.

\bibitem[LeCun et~al.(1998)LeCun, Bottou, Bengio, and
  Haffner]{lecun1998gradient}
LeCun, Yann, Bottou, L{\'e}on, Bengio, Yoshua, and Haffner, Patrick.
\newblock Gradient-based learning applied to document recognition.
\newblock \emph{Proceedings of the IEEE}, 1998.

\bibitem[Murray \& Salakhutdinov(2009)Murray and
  Salakhutdinov]{murray2009evaluating}
Murray, Iain and Salakhutdinov, Ruslan~R.
\newblock Evaluating probabilities under high-dimensional latent variable
  models.
\newblock In \emph{Advances in Neural Information Processing Systems}, 2009.

\bibitem[Neal(1992)]{neal1992connectionist}
Neal, Radford~M.
\newblock Connectionist learning of belief networks.
\newblock \emph{Artificial intelligence}, 1992.

\bibitem[Raiko et~al.(2014)Raiko, Li, Cho, and Bengio]{raiko2014iterative}
Raiko, Tapani, Li, Yao, Cho, Kyunghyun, and Bengio, Yoshua.
\newblock Iterative neural autoregressive distribution estimator {NADE}-k.
\newblock In \emph{Advances in Neural Information Processing Systems}, 2014.

\bibitem[Rezende et~al.(2014)Rezende, Mohamed, and
  Wierstra]{rezende2014stochastic}
Rezende, Danilo~J, Mohamed, Shakir, and Wierstra, Daan.
\newblock Stochastic backpropagation and approximate inference in deep
  generative models.
\newblock In \emph{Proceedings of the 31st International Conference on Machine
  Learning}, 2014.

\bibitem[Russakovsky et~al.(2015)Russakovsky, Deng, Su, Krause, Satheesh, Ma,
  Huang, Karpathy, Khosla, Bernstein, Berg, and Fei-Fei]{ILSVRC15}
Russakovsky, Olga, Deng, Jia, Su, Hao, Krause, Jonathan, Satheesh, Sanjeev, Ma,
  Sean, Huang, Zhiheng, Karpathy, Andrej, Khosla, Aditya, Bernstein, Michael,
  Berg, Alexander~C., and Fei-Fei, Li.
\newblock {ImageNet Large Scale Visual Recognition Challenge}.
\newblock \emph{International Journal of Computer Vision (IJCV)}, 2015.

\bibitem[Salakhutdinov \& Hinton(2009)Salakhutdinov and
  Hinton]{salakhutdinov2009deep}
Salakhutdinov, Ruslan and Hinton, Geoffrey~E.
\newblock Deep boltzmann machines.
\newblock In \emph{International Conference on Artificial Intelligence and
  Statistics}, 2009.

\bibitem[Salakhutdinov \& Murray(2008)Salakhutdinov and
  Murray]{salakhutdinov2008quantitative}
Salakhutdinov, Ruslan and Murray, Iain.
\newblock On the quantitative analysis of deep belief networks.
\newblock In \emph{Proceedings of the 25th international conference on Machine
  learning}, 2008.

\bibitem[Salimans et~al.(2015)Salimans, Kingma, and
  Welling]{salimans2014markov}
Salimans, Tim, Kingma, Diederik~P, and Welling, Max.
\newblock Markov chain monte carlo and variational inference: Bridging the gap.
\newblock \emph{Proceedings of the 32nd International Conference on Machine
  Learning}, 2015.

\bibitem[Sohl{-}Dickstein et~al.(2015)Sohl{-}Dickstein, Weiss, Maheswaranathan,
  and Ganguli]{deepdiffusion}
Sohl{-}Dickstein, Jascha, Weiss, Eric~A., Maheswaranathan, Niru, and Ganguli,
  Surya.
\newblock Deep unsupervised learning using nonequilibrium thermodynamics.
\newblock \emph{Proceedings of the 32nd International Conference on Machine
  Learning}, 2015.

\bibitem[Stollenga et~al.(2015)Stollenga, Byeon, Liwicki, and
  Schmidhuber]{NIPS2015_5642}
Stollenga, Marijn~F, Byeon, Wonmin, Liwicki, Marcus, and Schmidhuber, Juergen.
\newblock Parallel multi-dimensional lstm, with application to fast biomedical
  volumetric image segmentation.
\newblock In \emph{Advances in Neural Information Processing Systems 28}. 2015.

\bibitem[Sutskever et~al.(2011)Sutskever, Martens, and
  Hinton]{sutskever2011generating}
Sutskever, Ilya, Martens, James, and Hinton, Geoffrey~E.
\newblock Generating text with recurrent neural networks.
\newblock In \emph{Proceedings of the 28th International Conference on Machine
  Learning}, 2011.

\bibitem[Theis \& Bethge(2015)Theis and Bethge]{theis2015generative}
Theis, Lucas and Bethge, Matthias.
\newblock Generative image modeling using spatial {LSTM}s.
\newblock In \emph{Advances in Neural Information Processing Systems}, 2015.

\bibitem[Theis et~al.(2015)Theis, van~den Oord, and Bethge]{theis2015note}
Theis, Lucas, van~den Oord, A{\"a}ron, and Bethge, Matthias.
\newblock A note on the evaluation of generative models.
\newblock \emph{arXiv preprint arXiv:1511.01844}, 2015.

\bibitem[Uria et~al.(2013)Uria, Murray, and Larochelle]{uria2013rnade}
Uria, Benigno, Murray, Iain, and Larochelle, Hugo.
\newblock {RNADE}: The real-valued neural autoregressive density-estimator.
\newblock In \emph{Advances in Neural Information Processing Systems}, 2013.

\bibitem[Uria et~al.(2014)Uria, Murray, and Larochelle]{uria2013deep}
Uria, Benigno, Murray, Iain, and Larochelle, Hugo.
\newblock A deep and tractable density estimator.
\newblock In \emph{Proceedings of the 31st International Conference on Machine
  Learning}, 2014.

\bibitem[van~den Oord \& Schrauwen(2014{\natexlab{a}})van~den Oord and
  Schrauwen]{van2014factoring}
van~den Oord, A{\"a}ron and Schrauwen, Benjamin.
\newblock Factoring variations in natural images with deep gaussian mixture
  models.
\newblock In \emph{Advances in Neural Information Processing Systems},
  2014{\natexlab{a}}.

\bibitem[van~den Oord \& Schrauwen(2014{\natexlab{b}})van~den Oord and
  Schrauwen]{van2014student}
van~den Oord, A{\"a}ron and Schrauwen, Benjamin.
\newblock The student-t mixture as a natural image patch prior with application
  to image compression.
\newblock \emph{The Journal of Machine Learning Research}, 2014{\natexlab{b}}.

\bibitem[Zhang et~al.(2016)Zhang, Chen, Yu, Yao, Khudanpur, and
  Glass]{zhang2016highway}
Zhang, Yu, Chen, Guoguo, Yu, Dong, Yao, Kaisheng, Khudanpur, Sanjeev, and
  Glass, James.
\newblock Highway long short-term memory {RNN}s for distant speech recognition.
\newblock In \emph{Proceedings of the International Conference on Acoustics,
  Speech and Signal Processing}, 2016.

\end{thebibliography}
\bibliographystyle{icml2016}


\begin{figure*}[p]

\centering
\includegraphics[trim={0 0 0 0},clip, width=0.95\linewidth]{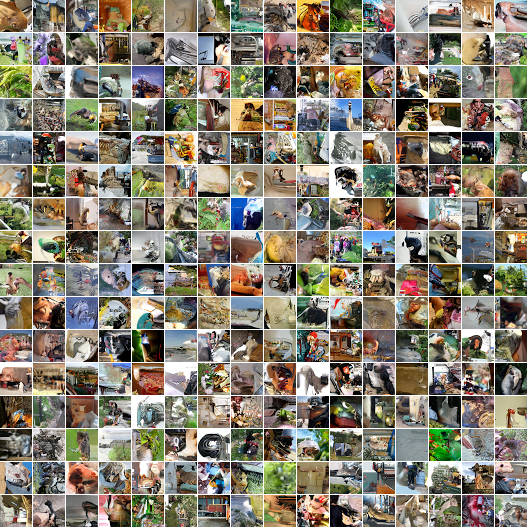}
\caption{Additional samples from a model trained on ImageNet 32x32 (right) images.}
\label{fig:appendix_imagenet_32}
\end{figure*}

\end{document}